\documentclass[11pt,a4paper]{article}
\usepackage[hyperref]{acl}
\usepackage{times}
\usepackage{latexsym}
\usepackage{amsmath}
\usepackage{amsfonts}
\usepackage{graphicx}
\usepackage{array}
\usepackage{longtable}
\usepackage{subcaption}
\usepackage{bbm}
\usepackage{booktabs} 
\usepackage{multirow}

\newcolumntype{L}[1]{>{\raggedright\let\newline\\\arraybackslash\hspace{0pt}}m{#1}}
\newcolumntype{C}[1]{>{\centering\let\newline\\\arraybackslash\hspace{0pt}}m{#1}}
\newcolumntype{R}[1]{>{\raggedleft\let\newline\\\arraybackslash\hspace{0pt}}m{#1}}

\usepackage{microtype}

\title{Open Vocabulary Extreme Classification Using Generative Models}

\author{
{\bf Daniel Simig},
{\bf Fabio Petroni},
{\bf Pouya Yanki},
{\bf Kashyap Popat},
{\bf Christina Du}, \\
{\bf Sebastian Riedel},
{\bf Majid Yazdani} \\ \\
Meta AI \\
\texttt{\{danielsimig,fabiopetroni,pya,kpopat,xiaodu,} \\ \texttt{sriedel,myazdani\}@fb.com}
}

\date{}

\newcounter{srCounter}
\newif\ifsrvar
\srvartrue
\ifsrvar
\newcommand{\seb}[1]{{\small \color{red} \refstepcounter{srCounter}\textsf{[SR]$_{\arabic{srCounter}}$:{#1}}}}
\else
\newcommand{\seb}[1]{}
\fi

\newcounter{fpCounter}
\newif\iffpvar
\fpvartrue
\iffpvar
\newcommand{\fabio}[1]{{\small \color{blue} \refstepcounter{fpCounter}\textsf{[FP]$_{\arabic{fpCounter}}$:{#1}}}}
\else
\newcommand{\fabio}[1]{}
\fi

\newcounter{dsCounter}
\newif\ifdsvar
\dsvartrue
\ifdsvar
\newcommand{\daniel}[1]{{\small \color{purple} \refstepcounter{dsCounter}\textsf{[DS]$_{\arabic{dsCounter}}$:{#1}}}}
\else
\newcommand{\daniel}[1]{}
\fi

\newcounter{kpCounter}
\newif\ifdsvar
\dsvartrue
\ifdsvar
\newcommand{\kashyap}[1]{{\small \color{brown} \refstepcounter{kpCounter}\textsf{[KP]$_{\arabic{kpCounter}}$:{#1}}}}
\else
\newcommand{\kashyap}[1]{}
\fi

\newcounter{apCounter}
\newif\ifapvar
\apvartrue
\ifapvar
\newcommand{\piktus}[1]{{\small \color{orange} \refstepcounter{apCounter}\textsf{[AP]$_{\arabic{apCounter}}$:{#1}}}}
\else
\newcommand{\piktus}[1]{}
\fi

\newcounter{plCounter}
\newif\ifplvar
\plvartrue
\ifplvar
\newcommand{\patrick}[1]{{\small \color{magenta} \refstepcounter{plCounter}\textsf{[PL]$_{\arabic{plCounter}}$:{#1}}}}
\else
\newcommand{\patrick}[1]{}
\fi

\newcounter{afCounter}
\newif\ifafvar
\afvartrue
\ifafvar
\newcommand{\angela}[1]{{\small \color{olive} \refstepcounter{afCounter}\textsf{[AF]$_{\arabic{afCounter}}$:{#1}}}}
\else
\newcommand{\angela}[1]{}
\fi

\newcounter{jtCounter}
\newif\ifjtvar
\jtvartrue
\ifjtvar
\newcommand{\jt}[1]{{\small \color{purple} \refstepcounter{jtCounter}\textsf{[JT]$_{\arabic{jtCounter}}$:{#1}}}}
\else
\newcommand{\jt}[1]{}
\fi

\def\model{\textsc{groov}}
\newcommand{\eg}{\textit{e.g.}}
\newcommand{\ie}{\textit{i.e.}}

\begin{document}
\maketitle
\begin{abstract}
The extreme multi-label classification~(XMC) task aims at tagging content with a subset of labels from an extremely large label set. The label vocabulary is typically defined in advance by domain experts and assumed to capture all necessary tags. However in real world scenarios this label set, although large, is often incomplete and experts frequently need to refine it. To develop systems that simplify this process, we introduce the task of open vocabulary XMC~(OXMC): given a piece of content, predict a set of labels, some of which may be outside of the known tag set. Hence, in addition to not having training data for some labels--as is the case in zero-shot classification--models need to invent some labels on-the-fly. We propose \model, a fine-tuned seq2seq model for OXMC that generates the set of labels as a flat sequence and is trained using a novel loss independent of predicted label order.
We show the efficacy of the approach, experimenting with popular XMC datasets for which \model~is  able to predict meaningful labels outside the given vocabulary while performing on par with state-of-the-art solutions for known labels.

\end{abstract}

\section{Introduction}

Extreme multi-label classification (XMC) aims at predicting a set of labels for a given input instance from an extremely large labels set ~\cite{pmlr-v48-yenb16,PPDsparse,BabSch17,BabSch19}. Examples for applying extreme classification are labeling a new article with Wikipedia's categories, classifying a product with catalog labels, classifying a resume into a collection of pertinent job titles. 


Despite the the scale of the label space, it is challenging to a priori capture all the possible ways in which an input instance can be categorized, especially at the industrial scale. Real-world e-Commerce platforms, for instance, contain billions of products from thousands of different categories that are continuously updated by human curators for all sort of reasons (\eg, see category changes in \citet{ebay}).

In this work we introduce the open vocabulary XMC task, where we measure the ability of models to go beyond the given vocabulary and automatically propose new labels that might complement the existing ones and fill gaps in the vocabulary. Note that this differs from a zero-shot formulation of the XMC problem \cite{Gupta21} where, although no training instance is available for some labels, they are still present in the given vocabulary.

To tackle the problem we propose \model, an autoregressive model that maps input sequences to a set of sequences. Inputs are documents/text, and outputs are collections of textual labels from an open vocabulary. We investigate multiple sequence-to-set-of-sequences instantiations, in particular an off-the shelf approach based on a encoder-decoder language model (T5, \citet{Raffel2019ExploringTransformer}) and a variant that uses a modified softmax function (\ie, multi-softmax) that does not penalize the model for assigning high probability to any gold label. This latter version inherently treats the target as a set of sequences (instead of a flat sequence) and outperforms the off-the shelf approach. 

To evaluate the out-of-vocabulary behaviour, we use popular XMC datasets for which a portion of the test labels do not appear in the train set. Differently from previous works, we assume models are unaware of such labels (\ie, they don't appear in the given label vocabulary) and need to find them with open-ended text generation. We show that \model~can indeed generate some of these labels while being competitive with state-of-the-art results on in-vocabulary metrics. 




In summary, the key contributions of this work are as follows: 
\begin{itemize}
\item introduce the open vocabulary  XMC task, where models are requested to classify content with meaningful labels that might not be present in the given vocabulary;
\item propose \model, a sequence-to-set-of-sequences model that can tag textual content with a set of labels from an open vocabulary;
\item present extensive analysis on the out-of-vocabulary behaviour of \model, including a human review of the generated labels.
\end{itemize}

\section{Related Work}  \label{section:related_work}
Traditionally Extreme Multi-Label Classification is done by the one-vs-all method. One-vs-all methods such as DiSMEC~\cite{BabSch17}, ProXML~\cite{BabSch19}, PDSparse~\cite{pmlr-v48-yenb16}, and PPDSparse~\cite{PPDsparse}, which treat each label as a binary classification problem, can achieve acceptable performance. One-vs-all methods suffer from computationally expensive complexity and large model size. Also, the classification tasks are independent of each other, and label dependency is not directly modeled. 
The high computational complexity in one-vs-all methods can be further improved by incorporating different partitioning techniques on the label spaces. For instance, Parabel~\cite{parabel} partitions the labels through a balanced 2-means label tree using label features constructed from the instances. Recently, several approaches have been proposed to improve Parabel. Bonsai~\cite{Bonsai} relaxes two main constraints in Parabel; allowing multi-way instead of binary partitionings of the label set at each intermediate node and also removing strict balancing constraints on the partitions. SLICE~\cite{slice} considers building an approximate nearest neighbor (ANN) graph as an indexing structure over the labels. For a given instance, the relevant labels can be found quickly from the nearest neighbors of the instance via the ANN graph. These models rely on sparse features engineered from the text, which is cumbersome and, most importantly, doesn't benefit from the knowledge of pre-trained LMs. Moreover, the partitioning of the label space is done independently from the classifier's training. In this paper, we leverage pre-trained language models and show that generative models efficiently partition the label space, token by token, and there is no need for crafting a tree of labels separate from the classifier. 

Deep learning models have improved extreme multi-label classification by learning better text representation from raw text. But the main challenge to these methods is how to couple with millions of labels. 
AttentionXML~\cite{attentionxml} shows success in extreme multi-label classification, overpassed all traditional machine learning methods, and proved the superiority of the raw text compared to sparse features. AttentionXML uses a label tree, and a new classification model is trained for each layer of this tree that makes inference slow in predicting. 
X-Transformer~\cite{chang2020taming} only uses pre-trained LMs to match the label clusters for a given raw text and then ranks these labels by linear classifications with the sparse features. X-Transformer is the first method of using pre-trained LMs in extreme multi-label classification. Due to the high computational complexity of transformer models, it only fine-tunes transformer models as the label clusters matcher, which can not fully exploit the power of transformer models.

Recently, GENRE~\cite{decao2020autoregressive} showed that seq2seq auto-recursive models using pre-trained models could effectively partition and traverse a set of large labels by generating tokens incrementally. In extreme multi-label classification, the output is a \textbf{set} of labels. Turning the set to a sequence of labels requires an ordering among labels, which might not be straightforward in many applications. \cite{vinyals2016order} shows that this ordering can significantly impact the performance. Authors in \cite{yang2018deep} propose an RL-based approach to relax the need for a fixed ordering among labels. We propose using a multi-softmax to relax the need for a fixed ordering which is much easier to train and implement than RL algorithms. Another advantage of our work to other set-output methods is that we model the multi-label classification as a set of sequences of tokens instead of a set of label identifiers. Therefore, we leverage more effectively the LM's knowledge in understanding each label.
\cite{Gupta21} tackles the problem of zero-shot learning in extreme multi-label classification in which it tags each input with a set of labels consisting of both seen and unseen labels during the training. Not only do we build an effective and efficient zero to few-shot learning, but we also want to go beyond that and tackle the problem of open vocabulary classification in which the taxonomy is not known to us entirely. 

Related to the open vocabulary extreme classification is the Open Set Recognition\cite{Geng2021RecentAI} in the computer vision community. Models proposed to solve the open set recognition have a different signature from our work. They define novel classes only in terms of sets of data points and do not generate names for classes that could then be compared against the true labels in a test set. Also, they operate only on images, and the methods' generalization to other modalities is not examined. Similar in spirit, \cite{hashtag} generates hashtags for microblogs and measures the ability of their model in generating new hashtags. The authors use a GRU-based dual encoder to generate hashtags. While there are similarities, our work is first in studying large generative pre-trained LM for open vocabulary extreme tagging by jointly modeling all golden labels using a novel loss (multi-softmax).
\section{Open-Vocabulary Tagging}

\newcommand{\tok}{\ensuremath{\mathrm{tok}}}

Consider N training data points $\{(X_i, Y_i)\}_{i=1..N}$ where $X_i$ is the text corresponding to the i-th instance and $Y_i \subseteq Y^*$ is the set of tags $X_i$ was annotated with. Importantly, we consider the set of all possible tags  $Y^*$ to be unknown both at training and inference time. We do assume, however, that each tag $l_k \in Y^*$ can be described by natural language, that is by a sequence of tokens, $\tok(l_k) = \{t_{k, j}\}_{j=1..len(l_k)}$. Lastly let  $Y_{seen} = \bigcup\limits_{i=1}^N Y_i$ denote the set of labels encountered at training time. Throughout this work we will pay special attention to labels outside of this set, which we refer to as unseen labels.

The above presented formulation of the topic tagging task is incompatible with currently prevalent XMC paradigms in several ways:
First, most traditional classifiers require not only $Y^*$ to be known in advance, but assume that for each label $l_k$ there are some examples tagged with $l_k$ so that a classifier can be learned for that particular label. These methods often don't rely on the label representation  $\tok(l_k)$ itself.
Second, more recent zero-shot work \citep{Gupta21} makes tagging possible even for previously unencountered labels $y_k \not\in Y_{seen}$. To our best knowledge, all of these methods rely on access to $Y^*$ in order to build some kind of index using label features.
Finally, current datasets have their limitations too: \cite{JainExtremeApplications} and \cite{SchultheisUnbiasedLabels} highlight that as the set of possible label grows it is unrealistic to expect that human annotators consider every single possible label in $ Y^*$ when annotating a document, thus we can expect all extreme classification datasets to be generally under-annotated. As we will see in Section \ref{subsection:novel_analysis} this hinders our ability to measure the precision of any open vocabulary tagging system.

In the following section we introduce a novel class of models that is particularly well-suited for exploring the whole label space $Y^*$ while maintaining good performance on the set of known labels $Y_{seen}$.

\section{Model}  \label{section:model}
Below we illustrate how to frame OXMC as seq2seq problem, propose a loss captures the set-nature of label sets more directly and then show how individual labels in the sets can be scored.  

\newcommand{\Concat}{\ensuremath{\mathrm{Concat}}}

\subsection{Seq2Seq for Sets of Sequences}

Given input text X, and some already produced output tokens $y_1, .., y_{i-1}$, \textit{seq2seq} models predict the probability of the next token in the output: $p(y_i|X, y_1, .. , y_{i-1})$. 
Open-vocabulary topic tagging can also be formulated as such sequence-to-sequence problem: Given text $X_i$, a set of tags $Y_i$ and a permutation $\pi$ that returns an ordered list of the elements of $Y_i$, we ask the model to predict the concatenation of the appropriate tags\footnote{In practice, we insert a special [SEP] token between each tag, making this mapping bijective which is required for decoding}  in the order defined by $\pi$. Formally, the target output can be defined as 
\[
T(Y_i, \pi) = \Concat\Bigl(\Bigl[\tok(\pi(Y_i)[k])\Bigr]_{k=1}^{|Y_i|}\Bigr).
\]

The need for the extra permutation input  $\pi$ in T reflects the fact that we are attempting to use a \emph{sequential} model that produces ordered list of tokens to predict an \emph{unordered set} of labels. This has a number of practical implications that we need to address.
At training time one needs to decide which ordering of the labels to feed to the model as target.
At inference time, the model might assign different probabilities to different orderings of the very same set of labels (as opposed to traditional classifiers that would assign a well defined probability to a particular set of labels)

\paragraph{Training}
During finetuning, for each training example, we uniformly sample a random permutation $\pi$ of the gold labels. Formally, this method corresponds to a loss function described in Equation \ref{fig:expectation_loss} 
\begin{equation}
\begin{aligned}
    \mathcal {L}(\theta)     & =  -\mathop{\mathbb{E}}_{\pi}\left[\log\Bigg( P_{\pi}(Y_i|X_i, \theta) \Bigg)\right] \\
    P_{\pi}(Y_i|X_i, \theta) & = \prod_{k=1}^{|Y_i|} P\Bigl(T[k] \bigg|T[1:k-1], X, \theta\Bigr) \\
                           T & = T(Y_i, \pi)
\end{aligned}
\label{fig:expectation_loss}
\end{equation}
\paragraph{Inference}
At inference time we decode the model naively choosing the most likely next token at each decoding step. We then split the produced output text by the separator token, resulting in a set of strings - these will be our predicted tags. Note that there’s no guarantee that the tags generated this way will be part of the labels used in the dataset, but our hope is that the model will learn what constitutes a good tag. For the purpose of computing position-based metrics such as Precision@K, PSP@K, NLSR@K we use the order in which the model produced the labels.



We refer to our training and inference approach as GROOV (GeneRative Out-Of-Vocabulary) tagging.

\subsection{Multi-Softmax Loss}

Assume a training example has gold labels A, B, and C and that in a particular training step we feed the permutation B, A, C to the model as the target. Let the logit corresponding to the first tokens of labels A, B, C be $z_A, z_B, z_C$. The softmax function inside the Cross-Entropy loss will be as follows:

\begin{equation}
\sigma_B(z) = \frac{e^{z_B}}{\sum_{i=1}^{N}{e^{z_i}}}
\end{equation}

The sum in the denominator also includes terms for the logits $z_A, z_C$ and thus the loss will eventually increase if the model assigns higher probabilities to tokens corresponding to labels A and C - even though those predictions would be completely reasonable. Unfortunately, the more labels an example has on average, the more prevalent this problem will become. 

In order to overcome this issue, we propose a modified softmax function dubbed \emph{Multi-Softmax}~(MSM) that does not penalize the model for assigning a high probability to any token that could lead to decoding a gold label that has been not produced yet. At a given decoding step let G be the set of token indices that could lead to producing a gold label (in our example A, B or C). Then the multi-softmax function is defined as:

\begin{equation}
\sigma_G(z) = \frac{\sum_{i \in G}{e^{z_i}}}{\sum_{i=1}^{N}{e^{z_i}}}
\end{equation}

We experiment with replacing the softmax term in the Cross-Entropy loss of T5 to this newly proposed version in the hope that it will learn more efficiently.

\subsection{Scoring Labels}
With the proposed sequential approach there is no simple way to compute a score for an individual label: at decoding time we can only access the probability of the next label \textit{given the previously decoded labels}. Instead, all we have is a binary decision whether the label appeared in the model output or not. In real life applications this can be problematic as one can not control the sensitivity of the model by thresholding the scores. It also makes the model perform suboptimally on metrics (e.g. P@K) where the ordering of labels w.r.t their probability is crucial.

In order to compute a robust score for a given label, one might compute its marginal probability over all possible output sequences. Of course this is computationally intractable, so instead in practice we can run a beam search of beam size B and sum up the probabilities of the beams that contain a particular label in order to approximate its marginal probability. Let $b_1, .., b_B$ be the label sequences resulting from such a beam search. Our approximation to the marginal probability of label $l_i$ can be written as:

\begin{equation}
\begin{aligned}
P(l_i) = \sum_{k=1}^B{\mathbbm{1}(l_i \in b_k) P(b_k)}
\end{aligned}
\end{equation}

\section{Experimental Setting}

\subsection{Datasets}

In order to focus on the ability of models to tag text with previously unseen labels, one might consider using the same datasets that are used to benchmark traditional zero-shot XMC. We evaluate our models on the two topic tagging datasets\footnote{Other datasets in that work are focused on item similarity-based recommendation rather than real tagging.}~\cite{Gupta21} report results on. EURLex-4.3K~\cite{Chalkidis2019Large-ScaleLegislation} is a collection of roughly 50K EU Legal documents annotated with ~4.3K tags. Wikipedia-1M~\cite{Gupta21} is a large collection of Wikipedia articles associated with 1M+ Wiki categories.

The above two datasets all contain some amount of unseen labels (see Table \ref{data_stats}) but are on the two extreme sides of the spectrum:  EURLex-4.3K only contains 163 unseen labels, whereas most of the labels in the test set of Wikipedia-1M are in fact not present in the training set. In order to effectively study the open-vocabulary tagging properties of this new class of models, we construct a third dataset motivated by a real world scenario that aims to be in the middle of this spectrum.

The AmazonCat13K dataset introduced by \citet{McauleyHiddenText} contains descriptions of products on Amazon and categories in the product taxonomy associated with them. This dataset does not contain unseen labels in its test set, so we create a new dataset by 1) randomly choosing 1000 labels from the set of labels that appear in the training split and 2) moving all examples in the training set that contain any of these 1000 labels to the test set.\footnote{Due to strong correlations between labels, some labels outside of the original 1000 also disappear from the training set and becomes an unseen label} We refer to this newly introduced version of the AmazonCat13K dataset as AmazonCat-OV, as it enables measuring the Open Vocabulary performance of models.

\begin{table}[h!]
\small
\begin{center}
\resizebox{\columnwidth}{!}{%
\begin{tabular}{C{2.3cm}C{1cm}C{0.8cm}C{1.3cm}C{1.3cm}}
\toprule
\bf Dataset name &  $N_{train}$  &  $N_{test}$ & $|Y_{seen}|$ & $|Y_{unseen}|$ \\
\midrule 
\midrule
EURLex-4.3K & 45K &  6K & 4,108 & 163 \\
AmazonCat-OV &  1.1M &  0.4M & 11,460 & 1,870 \\ 
Wikipedia-1M &  2.3M &  2.7M & 495,107 & 776,612 \\
\bottomrule
\end{tabular}
}
\end{center}
\caption{\label{data_stats} Basic statistics of datasets used in this work}
\end{table}



\subsection{Evaluation Metrics}

We expect two basic properties from the proposed new class of models:

\begin{itemize}
    \item Irrespective of the new labels, these models need to perform just as well as other XMC models on the overall dataset (including more frequent tags too).
    \item Additionally, we expect our proposed models to produce some of the labels that it has never seen and has no knowledge of - demonstrating some understanding of the structure of the label space and the ability to generalize beyond a predefined taxonomy.
\end{itemize}

\newcommand{\seen}{\ensuremath{\mathrm{seen}}}
\newcommand{\unseen}{\ensuremath{\mathrm{unseen}}}

To that end, we evaluate our models using the following metrics:

\textbf{Propensity-Scored Precision @ K (PSP@K)} is a variant of the commonly used Precision@K metric introduced by \citet{JainExtremeApplications} that assigns higher rewards for getting infrequent labels right (and by extrapolation, even higher reward for previously unseen labels). The scoring function is motivated by the observation that less frequent tags are more likely to be under-labeled as well as by the intuition that tagging with more granular, less frequent tags is likely of more value. We refer to the original paper for the implementation details of this metric. Code for computing this metric is provided by the Extreme Classification Repository \cite{Bhatia16}
    
\textbf{Metrics on unseen labels.} For a data point with model predictions $\widetilde Y_i$ and gold labels $Y_i$, let $Y_{\unseen, i} = Y_i \setminus Y_{\seen}$ and $\widetilde Y_{\unseen, i} = \widetilde Y_i \setminus Y_{\seen}$.

We calculate the standard Precision@K and Recall@K metrics considering these two sets, $\widetilde Y_{\unseen, k}$ and  $Y_{\unseen, k}$. On top of these instance-wise metrics we also define a metric on the entire test set that measures how many of the unseen labels in the test set has the model produced at least once. We call this the Novel Label Set Recall and define it as 
\[
NLSR@K = \frac{\Bigl|\bigcup\limits_{i=1}^{N_{test}} sorted(\widetilde Y_{\unseen, i})[:K]\Bigr|}{\Bigl|\bigcup\limits_{i=1}^{N_{test}}Y_{\unseen, i}\Bigr|}
\]
This formulation is motivated by potential future use cases of this novel family of models. One might run model inference on a new batch of data and collects the top-K out-of-vocabulary labels produced by the model from each data point. This set of novel tags could now be inspected manually and used to expand the taxonomy of labels if deemed appropriate. NLSR@K is an approximation for what percentage of the expansion of the label space could be captured by such a process.
    
\textbf{Soft-matching based metrics} Since the model has no knowledge of what the gold labels might look like, it is possible that it would produce some labels that are semantically equivalent to a gold label but would have a slightly different surface form. We investigate this and propose new metrics to address this in Section \ref{subsection:lexical_semantic_similarity}

\subsection{Training and Evaluation Setup}
Unless otherwise reported, we use the T5-base model obtained from Huggingface \cite{wolf-etal-2020-transformers}. We finetune these models on 4 Nvidia V100 GPUs using batch size 32 and AdamW optimizer with LR=0.0001. On datasets where validation set is not provided, we train for a fix number of Epochs (100 and 1 for EURLex and AmazonCat-OV respectively) and use beam size 15 for decoding. For the experiments on Wikipedia dataset, we train T5-base models for 3 epochs and T5-large model for 1 epoch, respectively. Beam size is set to 15 for decoding purpose.

\section{Quantitative Results}
First, this section looks at our model's overall performance on XMC, considering all the tags. Then, we look at the out of vocabulary performance by relaxing the definition of label matching to account for semantically similar labels with different surface forms. 
\subsection{Overall Performance}
Table \ref{psp_table} contains our results on entire label set, as measured by the PSP@K metric introduced above. Given the large number of XMC  models available today, we only show the top-few best performing models from each family of models that we referenced in Section \ref{section:related_work}. Note that all models \textit{except for our proposed models} have access to the overall set of labels at inference time.
Our simplest method that uses T5 as-is outperforms many of the XMC models developed in the past years. Using the methods described in Section \ref{section:model} we established a system that performs on par with the best available model on EUR-Lex4.3K and is the second-best model on Wikipedia-1M, only ~2\% point below the designed explicitly for the zero-shot model. No model outperforms our models on both datasets. Our scoring by marginalization improves the performance in \textbf{Wikipedia-1M} dataset, especially at the top 3 and 5 tags, showing it effectively builds a calibrated score for labels. But, in \textbf{EUR-Lex 4.3K}, the default order of the labels in the vanilla T5 model scores as high as ranking by marginalization. We conjecture the generative model learns to output the more confident tags first and then moves to the less confident ones. Our MultiSoftmax loss consistency improves the performance in comparison to the base model.
\begin{table*}[t]
\centering
\small
    \begin{tabular}{L{5.7cm}C{1.1cm}C{1.1cm}C{1.1cm}|C{1.1cm}C{1.1cm}C{1.1cm}}
     \toprule
     \multirow{2}{*}{\textbf{Algorithm}} & \multicolumn{3}{c|}{\textbf{EUR-Lex 4.3K}} & \multicolumn{3}{|c}{\textbf{Wikipedia-1M}} \\
     \cmidrule{2-4} \cmidrule{5-7}
      & \textbf{PSP@1} & \textbf{PSP@3} & \textbf{PSP@5} & \textbf{PSP@1} & \textbf{PSP@3} & \textbf{PSP@5} \\
     \midrule \midrule
     
     \model & 50.2 & 62.4 & 67.3 & 9.5 & 9.7 & 9.1 \\
     \midrule
     + sorted by marginal probabilities  & 50.2 & 62.4 & 67.3 & 9.6 & 13.2 & 15.6 \\
     + MSM  & 52.6 & \textbf{63.6} & 67.2 & 9.8 & 13.4 & 15.8 \\
     + T5-large & 52.6 & \textbf{63.6} & 67.7 & 10.1 & 13.1 & 15.2 \\
     \midrule
     ZestXML-tuned~\cite{Gupta21} & 48.01 & 60.29 & 66.15  & \textbf{14.43} & \textbf{15.80} & \textbf{17.31}\\
     AttentionXML~\cite{attentionxml}  & 53.92 & \textbf{63.59} & \textbf{67.85}  & 3.82 & 4.54 & 5.20 \\
     XReg~\cite{xreg} & \textbf{58.06} & 62.99 & 65.97 & 3.48 & 3.51 & 3.83 \\
     Parabel~\cite{parabel} & 46.82 & 58.8 & 64.29 & 2.99 & 3.32 & 3.65 \\ 
     DiSMEC~\cite{BabSch17} & 47.26 &  59.82 & 65.55 & 2.35 & 2.99 & 3.48 \\
     Bonsai~\cite{Bonsai} & 46.41 & 58.83 & 64.44 & 3.19 & 3.61& 4.05 \\
     PfastreXML~\cite{pfast} & 55.30 & 58.00 & 59.91&  2.97 & 2.90 & 3.10 \\
     
     \midrule
     
     FastText ANNS~\cite{fasttext}  & 17.10 & 15.74 & 16.13 & 7.16 & 6.01 & 6.19 \\
     BERT ANNS~\cite{bert-anns} & 4.64 & 3.66 & 3.57 & 10.34 & 8.17 & 8.20 \\

     \bottomrule
    \end{tabular}
    
\caption{\label{psp_table} PSP@K metrics on the full set of labels}
\end{table*}

\subsection{Out-Of-Vocabulary Performance}
What distinguishes our model from previous zero-shot approaches is that it is able to generate previously unseen labels without being told about their existence in advance. Table~\ref{oov_table} shows our measurements of recall and precision when only considering unseen labels. For this section, we use the two larger datasets with a reasonably large set of unseen labels. To our best knowledge \textit{no other XMC system can achieve a non-zero score in this setting}.
Recall@K metrics on both of these datasets demonstrate that the model can generalize beyond the labels it has seen and produce correct, novel labels in some percentage of the cases - although there is room for significant improvements still. 
A highlight is that on the AmazonCat-OV dataset, nearly one-quarter of the labels that we removed from the training set were generated as the top out-of-vocabulary prediction at least once in the test set.
Due to the ambiguous nature of evaluating open-vocabulary tags produced by generative models, recall and precision measurements based on exact label match are merely a lower bound on the practical performance of the model. We investigate this further in the following sections and find that these numbers are underestimating our model's true ability to produce previously unseen but valid tags.
\begin{table}[h!]


 


\begin{center}
\begin{tabular}{L{2.8cm}C{0.8cm}C{0.8cm}C{0.8cm}}
\toprule 
\bf AmazonCat-OV & \textbf{@1} &  \textbf{@3} &  \textbf{@5} \\
\midrule \midrule
Recall &  6.6 &  7.3 & 7.3 \\
Precision &  8.3 &  3.1 & 1.9 \\
NLSR &  23.9 &  25.8 & 25.9 \\
\end{tabular}
\end{center}

\begin{center}
\begin{tabular}{L{2.8cm}C{0.8cm}C{0.8cm}C{0.8cm}}
\toprule 
\bf Wikipedia-1M & \textbf{@1} &  \textbf{@3} & \textbf{@5} \\
\midrule \midrule
Recall &  3.2 &  9.4 & 13.3 \\
Precision &  5.4 &  5.4 & 4.7 \\
NLSR &  3.6 &  11.0 & 16.0 \\
\bottomrule
\end{tabular}
\end{center}
\caption{\label{oov_table} Performance of our best performing models on the set of unseen labels}
\end{table}







\subsection{Lexical/Semantic Similarity instead of Exact Matching} \label{subsection:lexical_semantic_similarity}
Some of the reasonable labels produced by the model may not exactly match the labels from the golden label set. This mismatch could be due to small lexical differences such as different spelling, hyphenation, pluralization, lexical form or capitalization. Additionally, the mismatch can be due to related terms or synonyms being generated instead of the exact label (for example \textit{"Kids' books"} instead of \textit{"Childrens' books"}).
Metrics like precision and recall would count all such generations as false positives, and this may not accurately describe the generative model's performance. To tackle this, we also measure soft precision and soft recall.
We introduce Soft Lexical Recall/Precision, which addresses the lexical differences. These metrics work exactly in the same way as normal precision and recall with the difference that any generated label $\hat{Y}$ is matched with a label from the golden set $Y$ if their edit distance is smaller than \(|\hat{Y}| / DF + 1\), where $DF$ is the division factor used to regulate the flexibility and accuracy of this matching. In our measurements we set $DF=10$.
We also introduce Soft Semantic Recall/Precision to address the problem with slightly different formulations of the same label or synonym words in the labels. Similar to the Soft Lexical metrics described above, we change the matching criteria between $\hat{Y}$ and $Y$ from exact lexical match to a BertScore \cite{zhang2020bertscore} based metric. We check the F1 score generated by BertScore and use a threshold of $0.94$ \footnote{BertScore Hash: roberta-large\_L17\_no-idf\_version=0.3.10(hug\_trans=4.12.3)\_fast-tokenizer}. This threshold is selected to make sure soft semantic matches correlates highly with sensibility in our human evaluation. This is shown in Table \ref{table:sensible_informative}.
Table~\ref{table:soft_metrics_table} shows the performance on the AmazonCat dataset using soft lexical and semantic matching alongside the exact precision and recall. The threshold in semantic and lexical matching is stringent; they highly correlate with sensibility in our human evaluation (e.g., 96\% in table \ref{table:sensible_informative}).  Still, we observe significant improvement in our precision/recall compared to the exact match, confirming that the model generates some correct tags with slight surface differences.

\begin{table}[h!]
\begin{center}
\begin{tabular}{C{1.7cm}C{0.9cm}C{0.9cm}C{0.9cm}}

\toprule
\multicolumn{4}{c}{\bf Recall Metrics} \\
\midrule
\bf Method & \textbf{@1} &  \textbf{@3} &  \textbf{@5} \\
\midrule
 \midrule
 Exact &  6.62 &  7.31 & 7.34 \\
 Lexical &  7.84 &  9.76 & 10.58 \\
 Semantic &  8.07 &  9.04 & 9.07 \\
\bottomrule
\multicolumn{4}{c}{\bf Precision Metrics} \\
\midrule
\textbf{Method} & \textbf{@1} &  \textbf{@3} &  \textbf{@5} \\
\midrule
 \midrule
 Exact &  8.34 &  3.13 & 1.89 \\
 Lexical &  9.83 &  4.17 & 2.71 \\
 Semantic &  10.21 &  4.34 & 2.65 \\
 \bottomrule
\end{tabular}

\end{center}
\caption{\label{table:soft_metrics_table} Precision/Recall of the model with exact matching as well as lexical and semantic soft matching on AmazonCat dataset}
\end{table}

\section{Human Review of Out of Vocabulary Generations} \label{subsection:novel_analysis}
\subsection{Interpreting Model Behavior}
In our experiment with the AmazonCat-OV dataset, our model correctly generated more than 400 different, novel categories that only appeared in the test set as ground truth labels. In order to qualitatively understand what type of model behavior led to producing these labels, we manually compared the input texts and the generated novel labels. 

We found that in most cases (89\%) the model effectively employs a very simple two-step strategy. First it identifies an n-gram in the input text that could be a meaningful category. Then the model decides if it makes sense to generate a label that is the verbatim copy of this n-gram ("London", "Table Tennis", "Bartending") or alternatively, it converts the n-gram into its plural form ("Kitchen Sinks", "Sleeping Pads").

In the rest of the cases (11\%), however, we found evidence that the model is able to creatively \textit{compose information} from across the item description in order to produce a label that does not appear verbatim in the text. Some examples of these labels are: "Wine Glasses", "Baby Food", "Patio Furniture Sets", "Lens Accessories".

\subsection{Sensibleness and Informativeness of Novel Labels}
Sometimes the model generates completely new terms that do not appear as a ground truth label in the test set. Even though these could indeed be false positives - as no taxonomy is ever complete - they could also be \textit{sensible}, and \textit{informative} new tags that could help the taxonomists expand the known label set. Due to this, our quantitative precision results might significantly underestimate the usefulness of the generated novel labels.

We inspected a random sample of 100 model predictions (142 novel labels) containing out of vocabulary labels and manually assessed their sensibleness and informativeness using human review. This is similar to the work of \citet{shuster2021retrieval}, where Consistency, Engagingness, and Knowledgeability of the responses of generative models in a conversational setting were manually measured. We focus on the two characteristics of sensible and informative as a new tag in the taxonomy needs to be both. It needs to make sense while being different enough from existing labels.

\begin{figure}[h!]
\begin{subfigure}{1.\columnwidth}
\includegraphics[width=\linewidth]{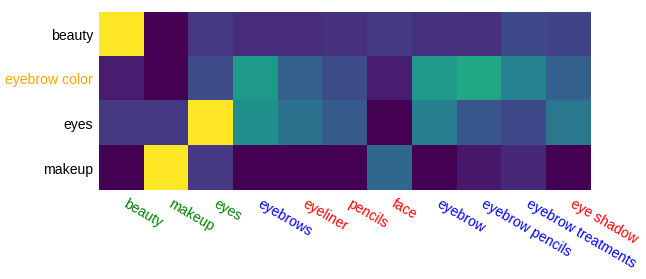}
\noindent\fbox{%
    \parbox{\linewidth}{%
        {\small \textbf{Input Text:} \input{appendix_a_novel_generations/8.txt}}
    }%
}
\caption{Sensible and Informative novel generated label}
\label{fig:good_example_novel_label}
\end{subfigure}
\begin{subfigure}{1.\columnwidth}
\includegraphics[width=\linewidth]{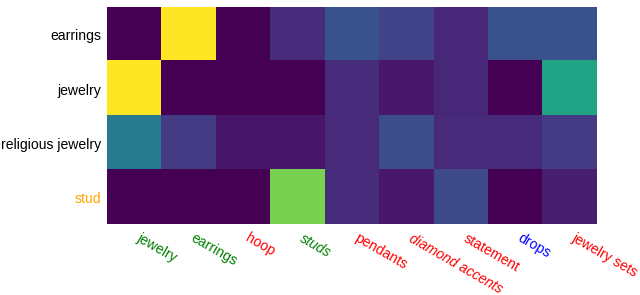}
\noindent\fbox{%
    \parbox{\linewidth}{%
        {\small \textbf{Input Text:} \input{appendix_a_novel_generations/1.txt}}
    }%
}
\caption{Not Sensible and not Informative novel generated label}
\label{fig:bad_example_novel_label}
\end{subfigure}
\caption{Showing examples generated by the model. Figure \ref{fig:good_example_novel_label} showing a sensible and informative prediction while the prediction in Figure  \ref{fig:bad_example_novel_label} is both \underline{not sensible} and \underline{not informative}}
\label{fig:examples_novel_label}
\end{figure}

In Figure \ref{fig:examples_novel_label} we present two examples of novel, entirely out of vocabulary generated labels. The color-map denotes the lexical similarity of generated predictions to the golden set, with gold meaning a perfect match and black being a complete mismatch.
For this lexical similarity we use the Levenshtein distance similar to section \ref{subsection:lexical_semantic_similarity} with $DF=10$. The Y-Axis of the color map corresponds to the golden set labels, and the individual labels in the golden set are colored gold when they are missing from the training set. The X-Axis represents the generated labels. The labels that are predicted correctly(potentially with soft lexical matching) are colored green. Those predicted falsely from the label set are colored red. The labels that could not be matched with any labels from the known label set are colored blue.
In Figure \ref{fig:good_example_novel_label} we see that the model generates several completely novel labels \textit{"eyebrow pencils"}, \textit{"eyebrow treatments"} and both singular and plural forms of \textit{"eyebrow"}. These labels better describe the input text; however, they are missing from both the label set and the golden set. Taxonomists could use such a prediction to improve the taxonomy and potentially the training dataset itself.
On the contrary, the novel label generated in Figure \ref{fig:bad_example_novel_label} is not related to the input text at all and is just a false positive.

Quantified results of manual review of a subsample of novel predictions by the model can be seen in Table \ref{table:sensible_informative}. $65\% $ of the novel generations in this subset are sensible. This means they can be safely used as labels/tags.
But more interestingly, $26\%$ of the novel generated labels we observed were both sensible and informative. These are typically more precise labels (more granular) for the input text than the golden set labels. This result is interesting as it provides a direct tool for taxonomists to expand/improve their taxonomy. By going over the $1-5\%$ of novel generated labels, they can find a lot of new sensible and informative labels.

We also want to measure the ability of the semantic soft matching introduced in section \ref{subsection:lexical_semantic_similarity} against the newly introduced \underline{sensitive} and informative framework. 
We see in Table \ref{table:sensible_informative} that using the semantic matching with the mentioned threshold detects with $96\%$ precision the \underline{sensibleness} and it also improves the precision for detecting \underline{informativeness}. Decreasing the threshold decreases the precision of detecting sensible tags. However, its recall is not very high, and if we wanted to find all the sensible and informative labels, we would still need to do human labeling.

Some more examples of these novel labels generated by the model and their evaluation based on the sensible and informative characteristics can be found in Appendix \ref{appendix:novel}.
Note that as this manual labeling process is expensive and time-consuming, our initial sample sets have been small. In the future, the novel generated labels can be studied more thoroughly from different aspects.

\begin{table}[h!]
\begin{center}
\begin{tabular}{L{1.6cm}C{1.5cm}C{1.2cm}C{1.2cm}}

\toprule 
\bf Semantic Match & \bf \# Labels & \bf Sen \% & \bf Inf \% \\
\midrule \midrule
 Yes & $26$ &  96 &  38 \\
 No & $116$ &  59 &  23 \\
 Total & $142$ &  65 &  26 \\
\bottomrule
\end{tabular}
\end{center}
\caption{\label{table:sensible_informative} Human Review of Novel Label Generations on a subset AmazonCat dataset}
\end{table}

\bibliography{ECG,daniel_mendeley_import}
\bibliographystyle{acl_natbib_2022}

\appendix
\clearpage
\onecolumn
\section{Appendix: Detailed Summary of Novel Generated Labels or Unseen Labels in Gold Set}
\label{appendix:novel}

In this appendix we list the subset of novel generated labels or instances with unseen labels in their gold set by the model that we studied in section \ref{subsection:novel_analysis}.
Table \ref{table:novel_labels_generated_sample} shows a subsample of predictions using our model on the Amazon dataset and Table \ref{table:novel_labels_generated_sample_wiki} shows the same for the Wiki dataset. The color-map denotes the lexical similarity of generated predictions to the golden set with gold meaning a perfect match and black being a complete mismatch. For this lexical similarity we use the Levenshtein distance similar to section \ref{subsection:lexical_semantic_similarity}. The Y-Axis of the color map corresponds to the golden set labels and the individual labels in the golden set are colored gold when they are missing from the training set. The X-Axis represents the generated labels. The labels that are predicted correctly are colored green, those predicted falsely from the label set are colored red and the labels that could not be matched with any labels from the known label set are colored blue. In the left column, we discuss each such novel generated label and evaluate it based on our sensible and informative framework.

\begin{longtable}[c]{| @{}p{0.2\linewidth}@{} | m{0.7\linewidth} |}
\caption{\label{table:novel_labels_generated_sample} A sample of predictions where the model generated novel labels on AmazonCat dataset} \\
 \hline
 \multicolumn{1}{|c|}{\bf Novel Labels} & \multicolumn{1}{|c|}{\bf Lexical Similarity Map \& Input Text}\\
 \hline
 \endfirsthead

 \hline
 \multicolumn{1}{|c|}{\bf Novel Labels} & \multicolumn{1}{|c|}{\bf Lexical Similarity Map \& Input Text}\\
 \hline
 \endhead

 \hline
 \endfoot

 \hline
 \hline\hline
 \endlastfoot
  \hline
 \begin{tabular}{p{0.7\linewidth}} \textit{"air intake kits"}: sensible but \underline{not informative} as there is another very similar label in gold set that could have been generated \\ \hline \textit{"intake system"}: sensible but \underline{not informative} \\ \end{tabular}  & \includegraphics[width=\linewidth]{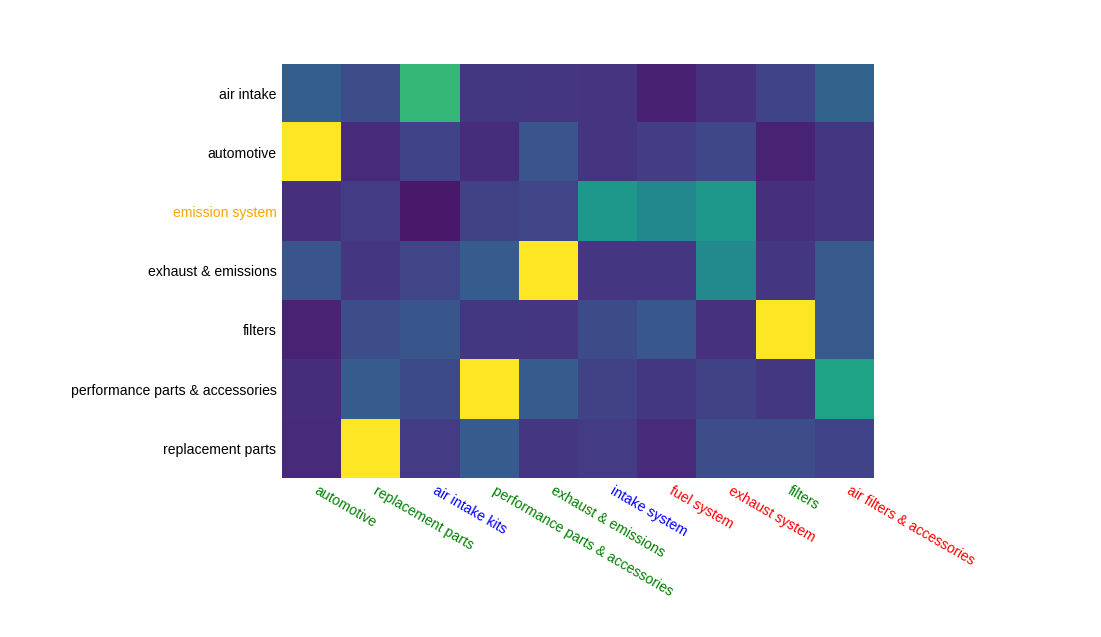} {\small K\&N 57-9014-1 Fuel Injection Performance Kit Gen2 Air Intake Kit The kit replaces your vehicle's restrictive factory air filter and air intake housing. K intake systems are designed to dramatically reduce intake restriction as they smooth and straighten air flow. This allows your vehicle's engine to inhale a larger volume of air than the OEM air filter assembly.  More air means more usable power and acceleration throughout the engine's RPM range.  The filters on these kits are washable, reusable and easy to install with tools commonly available.} \\
  \hline
  \begin{tabular}{p{0.7\linewidth}} \textit{"drops"}: \underline{Not sensible} \& \underline{not informative}\\ \end{tabular} & \includegraphics[width=\linewidth]{appendix_a_novel_generations/1.png} {\small 1/2 Carat Sterling Silver CZ Cross Stud Earrings The look of white gold at a silver price! These sterling silver earrings perfectly mimic white gold and diamonds with their rhodium finish and cubic-zirconia stones. Rhodium is a metal that is part of the platinum family. High-end silver and gold are rhodium treated to prevent oxidation and to have the white shiny look associated with platinum and white gold. These earrings' rhodium finish will prevent them from tarnishing.} \\
  \hline
  \begin{tabular}{p{0.7\linewidth}} \textit{"acoustic-electric basses"}: sensible and informative. This tag seems to be missing from label set and the closest matching ones \textit{"electric basses"} and \textit{"bass guitars"} is missing from golden set \\ \hline The other forms with \textit{"/"} and \textit{"and"} are similarly sensible and informative \\ \end{tabular} & \includegraphics[width=\linewidth]{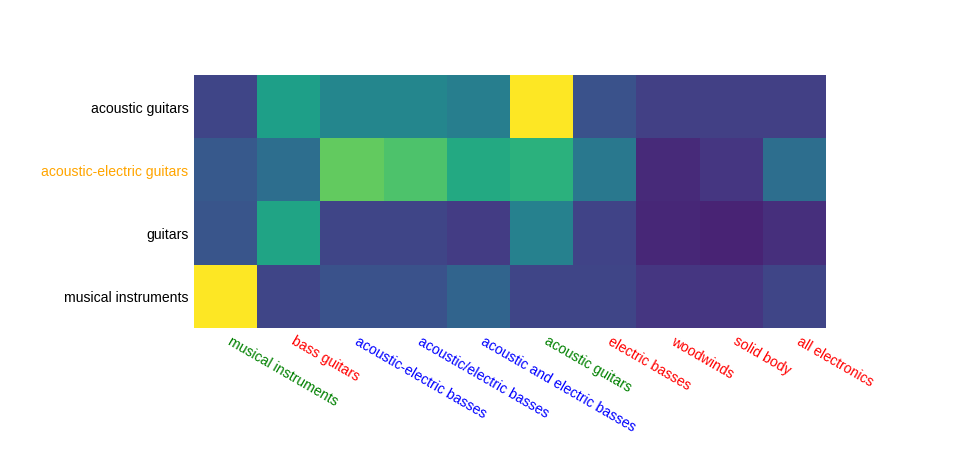} {\small Dean Acoustic-Electric Bass Cutaway Satin Finish Offering a large body with deep, full tone, this Dean acoustic-electric bass guitar (model EABC) also looks great on stage with a handsome satin-finished top made of select spruce wood and an abalone sound hole accent. It also features Dean's passive pre-amp electronics, a 34-inch scale, and a rosewood fingerboard with pearl dotted inlays.      Specifications      Top: Select spruce Body: Mahogany Neck: Mahogany Fingerboard: Rosewood with pearl dot inlays Bridge: Rosewood     Scale: 34 inches Tuners: Die cast Electronics: Dean passive pre-amp Finish: Satin natural	Dean EABC Electric Acoustic Bass is a Large Body, Big Sounding Acoustic Bass. Dean EABC comes with passive pre amp and is available in satin natural. Dean EABC is the BEST VALUE in a acoustic/electric bass on the market today. EABC ~Select Spruce Top ~34" scale ~Mahogany bound neck ~Rosewood fingerboard ~Pearl DOT Inlayes ~Die Cast Tuners ~Set Neck ~Celluliod Binding/Rosette ~R...} \\
  \hline
  \begin{tabular}{p{0.7\linewidth}} \textit{"ni-cad nails"}: \underline{Not Sensible} and \underline{not informative}. The input text is about nailers and not nails \\ \hline \textit{"straight nails"}: \underline{Not Sensible} and \underline{not informative} for similar reasons as above \\ \end{tabular} & \includegraphics[width=\linewidth]{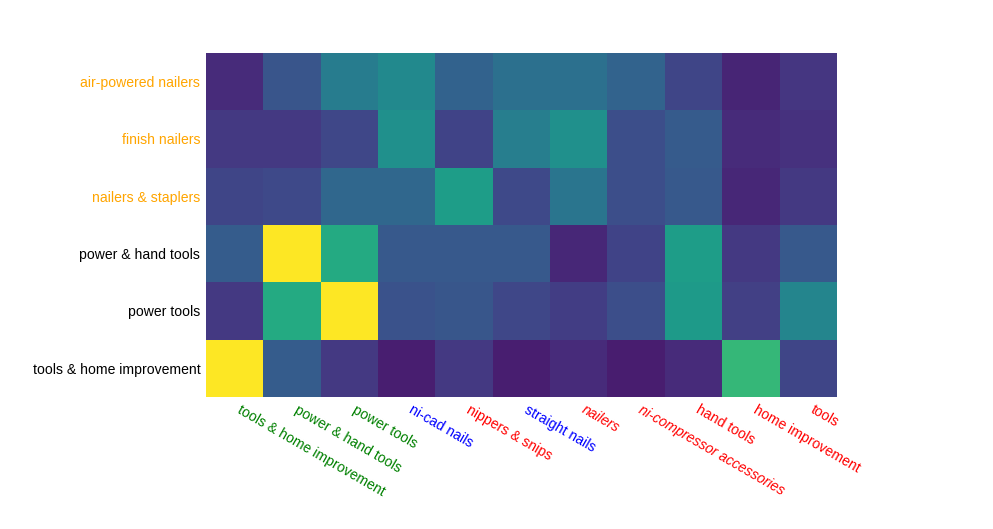} {\small DEWALT DC616KA 1-1/2-Inch to 2-1/2-Inch 18-Volt Ni-Cad Cordless 16-Gauge Straight Finish Nailer Kit No compressor. No hoses. No kidding. And no sacrifices in speed or power, either. There's absolutely no comparison between this performer and the fuel-cell powered competition, which we thought was a great innovation. But there's no costly fuel cell to replace on this tool-just pop on a recharged XRP battery and get back on the job. The only difference you'll feel between this and a traditional pneumatic is that you're not tethered to an air hose. It's just as fast and fires just as powerfully into both soft and hard joints. We love that you can choose bump or sequential mode for precision or speed, something most nailers don't offer, and the integrated headlight is another impressive addition, really lighting up your workpiece even in the worst conditions. There's a fantastic six-position numbered dial to reference your depths, so you can move easily between, for example, baseboard and ...} \\
  \hline
  \begin{tabular}{p{0.7\linewidth}} \textit{"usability"}: sensible and informative. The topic being discussed is Usability Inspection for UIs. The labels seems to be missing from both label set and golden set. \\ \end{tabular} & \includegraphics[width=\linewidth]{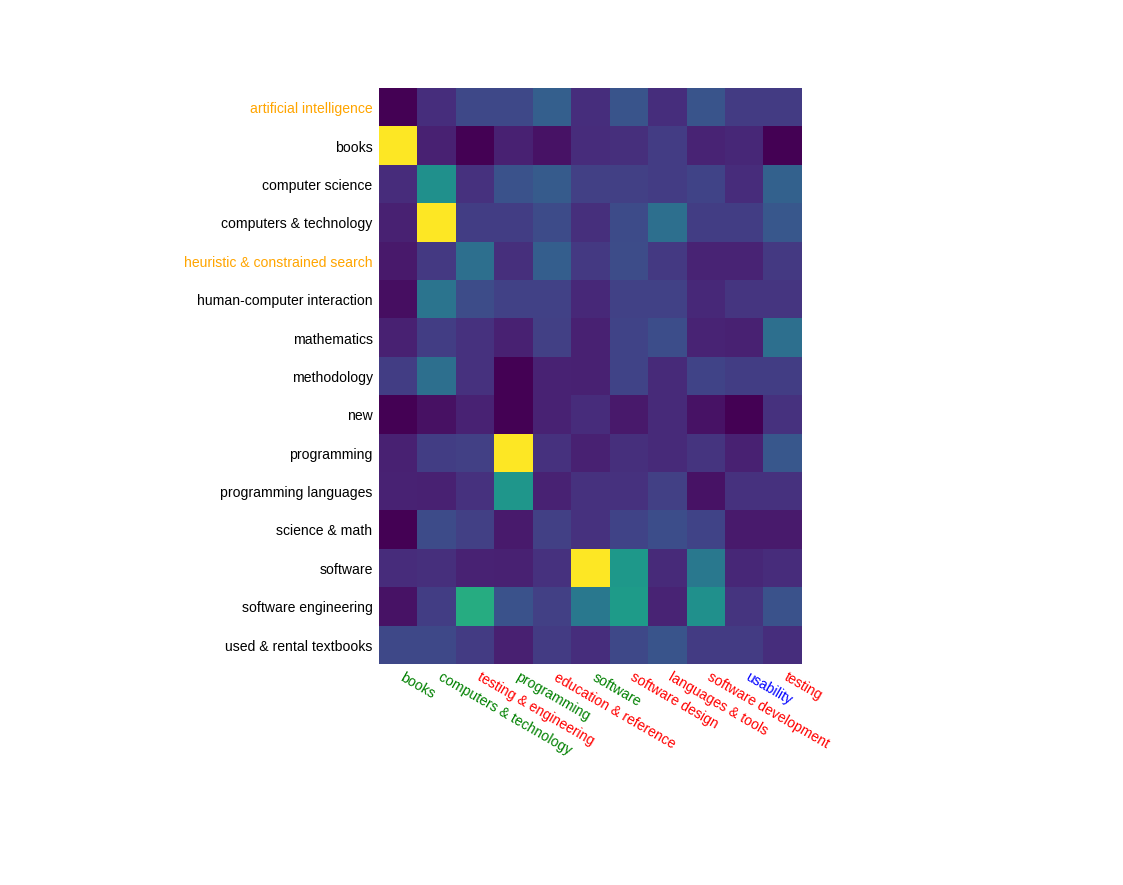} {\small Usability Inspection Methods Considered the founder of this research area, Nielsen presents a contributed exposition written by the foremost experts in this rapidly growing and important field. Devised for user interface practitioners searching for cost-effective ways of improving their designs, the book begins with descriptions of simple discount usability engineering methods such as heuristic evaluation which can be learned quickly and immediately applied to the reader's current project. Later chapters cover more formal inspection techniques offering additional benefits and discuss practical aspects of comparing the methods and user testing along with suggestions for when to use what techniques.	The last few years have seen the emergence of usability inspection (UI) as an important new tool to help user interface designers and software developers guarantee that their products meet the highest standards of usability. Everywhere UI methods have been implemented they have proven to be f...} \\
  \hline
  \begin{tabular}{p{0.7\linewidth}} \textit{"mono microphones"}: \underline{Not Sensible} and \underline{not informative} as mono microphones are not mentioned in text \\ \hline \textit{"single microphones"}: \underline{Not Sensible} and \underline{not informative} for similar reasons as above \\ \end{tabular} & \includegraphics[width=\linewidth]{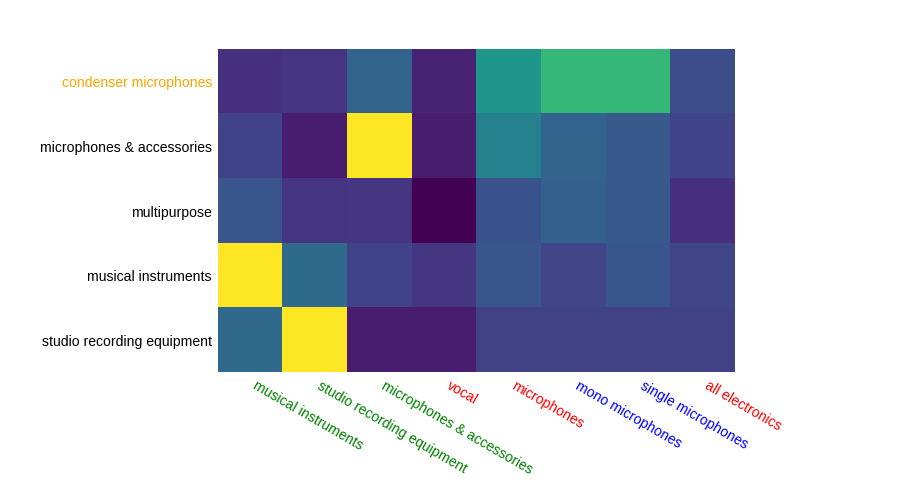} {\small Audio Technica ATM8010 ATM10a Artist Series Fixed-Charge 'Omni' Condenser Microphone Ideal for group vocals, strings, cymbal overheads, acoustic guitar and piano. Omni pattern provides maximum ambient pickup. Extremely smooth, extended response on- and off-axis. Low sensitivity to popping and overload. Operates on battery or phantom power.} \\
  \hline
  \begin{tabular}{p{0.7\linewidth}} \textit{"wrench holders"}: \underline{Not sensible} and \underline{not informative}. \\ \end{tabular} & \includegraphics[width=\linewidth]{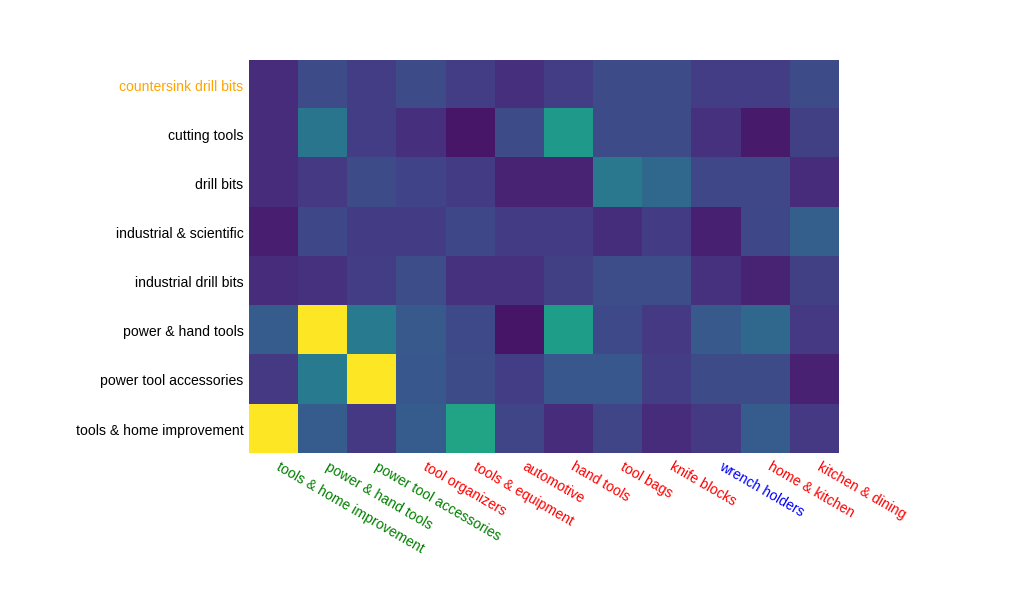} {\small DEWALT DW2050 Quick Change 3-Inch Magnetic Bit Tip Holder DeWalt DW2050 Quick Change 3-Inch Magnetic Bit Tip Holder	115-DW2050 Magnetic Holder Quick Change Magnetic Holder Unit Sold is in measure of 1 Box} \\
  \hline
  \begin{tabular}{p{0.7\linewidth}} \textit{"martini boxes"}: \underline{Not sensible} and \underline{not informative}. This mistake is perhaps due to the term \textit{"Martin"} being mentioned multiple times in another context in the input \\ \end{tabular} & \includegraphics[width=\linewidth]{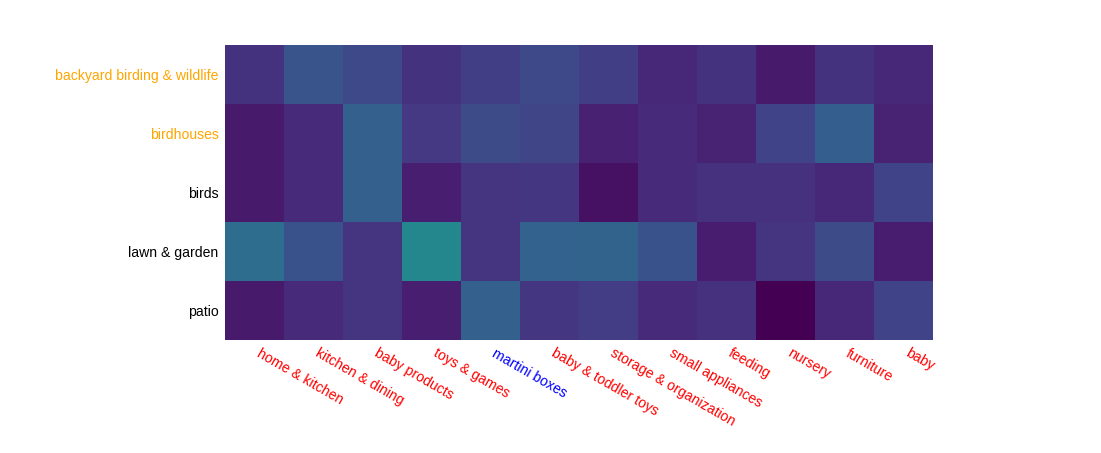} {\small Nature House M12K Trio Purple Martin Pioneer House Allow purple martins to colonize in your yard with the Trio Purple Martin Pioneer House. This home was one of the first ever built from aluminum, which helps keep the martins cool during the hot summer months. Such construction also offers durability to your martin house and will last several seasons. Each of the 12 compartments is 6 inches long x 6 inches wide x 6 inches high, the perfect size for martins, and has a 2.125 inch entrance hole.  Each compartment also has an individual lift up, snap out door so you can clean out one without disturbing the other nests. Guard rails along the porches of the home prevent babies from falling out of the nest and allow martins room to perch and preen. This is also accomplished with an included 22 inch roof perch. A set of 12 winter door stops close the house when your martins migrate south. The Pioneer home is compatible with any pole with a 1.25 inch outside diameter. Help purple martins nest i...} \\
\hline
\begin{tabular}{p{0.7\linewidth}} \textit{"eyebrow pencils"}: sensible and informative. This label describes the input text very precisely and the golden seems not to be complete. \\ \hline \textit{"eyebrow treatment"} \& \textit{"eyebrow"} sensible and informative like the above. \\ \end{tabular} & \includegraphics[width=\linewidth]{appendix_a_novel_generations/8.png} {\small NARS Eyebrow Pencil Sculpts and defines the eyebrow with rich, natural looking pigment to softly frame the eyes. The firm texture allows for maximum control and provides long-lasting definition.} \\
\hline
\begin{tabular}{p{0.7\linewidth}} \textit{"boot \& wheels"}: \underline{Not sensible} and \underline{not informative}. There seems to be a perfect label in the golden set that was also predicted \\ \end{tabular} & \includegraphics[width=\linewidth]{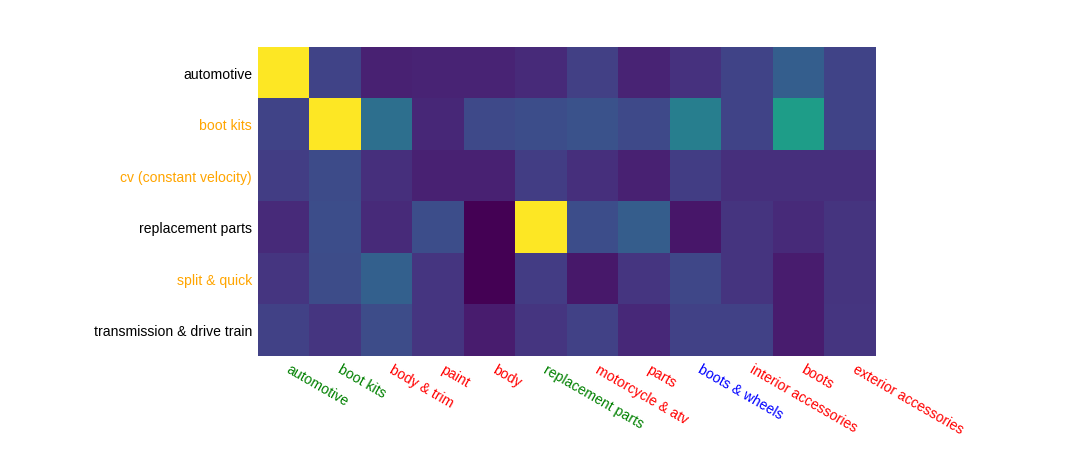} {\small Dorman 614-434 HELP! Constant Velocity Joint Quick Boot Kit Dorman Products, Inc. is well-known as a leader in providing quality auto parts to the aftermarket. We've earned our reputation for excellence from over three decades of experience in providing automotive replacement parts, fasteners and service line products primarily for the automotive aftermarket. Our prestigious position stems from a unique combination of application expertise, innovative product design, and breadth of product offerings, many of which are not conveniently or economically available elsewhere.  At Dorman, we take pride in the quality of our products and in your satisfaction.} \\
\hline
\begin{tabular}{p{0.7\linewidth}} \textit{"kids' books"}: sensible but \underline{not informative} as we have a similar known label \textit{"childrens' books"} \\ \end{tabular} & \includegraphics[width=\linewidth]{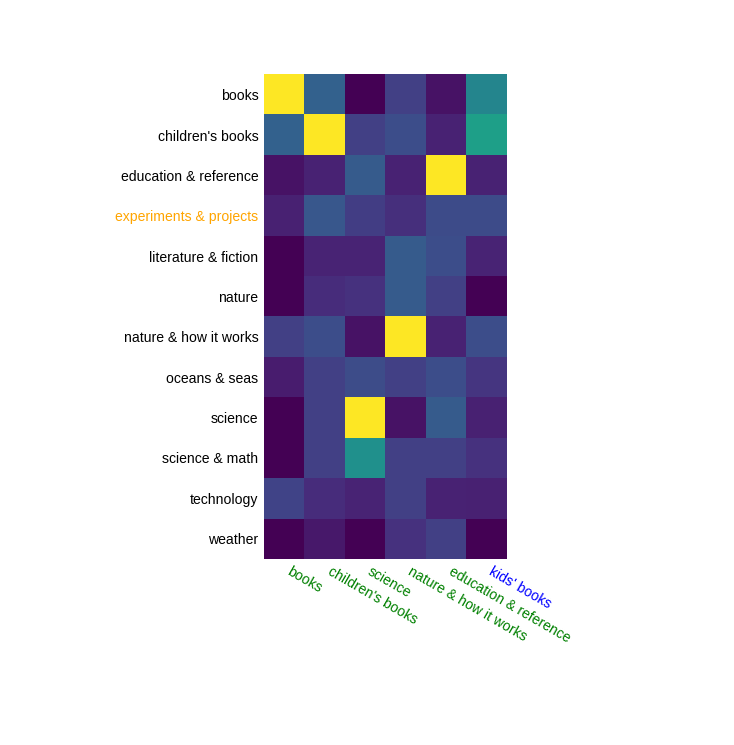} {\small Science in Seconds at the Beach: Exciting Experiments You Can Do in Ten Minutes or Less Science in Seconds at the Beach teaches children dozens of activities that investigate the mysteries of animals, plants, sand, shells, sun and water. Easy step-by-step instructions and illustrations are provided for each activity."--Asbury Park Press	Surf's up for science fun with these quick and easy activities. This book offers over 150 quick and easy experiments that will help children investigate the mysteries of animals, plants, sand, shells, sun, and water. Each activity takes ten minutes or less to complete, and answers a provocative question like: Do fish close their eyes? Can you hold your breath longer than a whale? How is sand made? How can seaweed forecast the weather? Do all snail shells coil in the same direction? And why do we seem to hear the ocean in empty sea shells?	Do fish close their eyes?Can you hold your breath longer than a whale?How is sand made?Why do we hear the ocean in e...} \\
\hline

\end{longtable}
\newpage

\begin{longtable}[c]{| @{}p{0.2\linewidth}@{} | m{0.7\linewidth} |}
\caption{\label{table:novel_labels_generated_sample_wiki} A sample of predictions where the model generated novel labels on Wiki dataset} \\
\hline
\multicolumn{1}{|c|}{\bf Novel Labels} & \multicolumn{1}{|c|}{\bf Lexical Similarity Map \& Input Text}\\
\hline
\endfirsthead

\hline
\multicolumn{1}{|c|}{\bf Novel Labels} & \multicolumn{1}{|c|}{\bf Lexical Similarity Map \& Input Text}\\
\hline
\endhead

\hline
\endfoot

\hline
\hline\hline
\endlastfoot
\hline
\hline
  
\begin{tabular}{p{0.7\linewidth}}
    \textit{"Events in the United States"}: sensible but \underline{not informative} \\
    \hline
    \textit{"Events in Washington DS"}: sensible but \underline{not informative} \\ 
    \textit{"Dinners in the United States"}: sensible but \underline{not informative}
\end{tabular} 
&
\includegraphics[width=\linewidth]{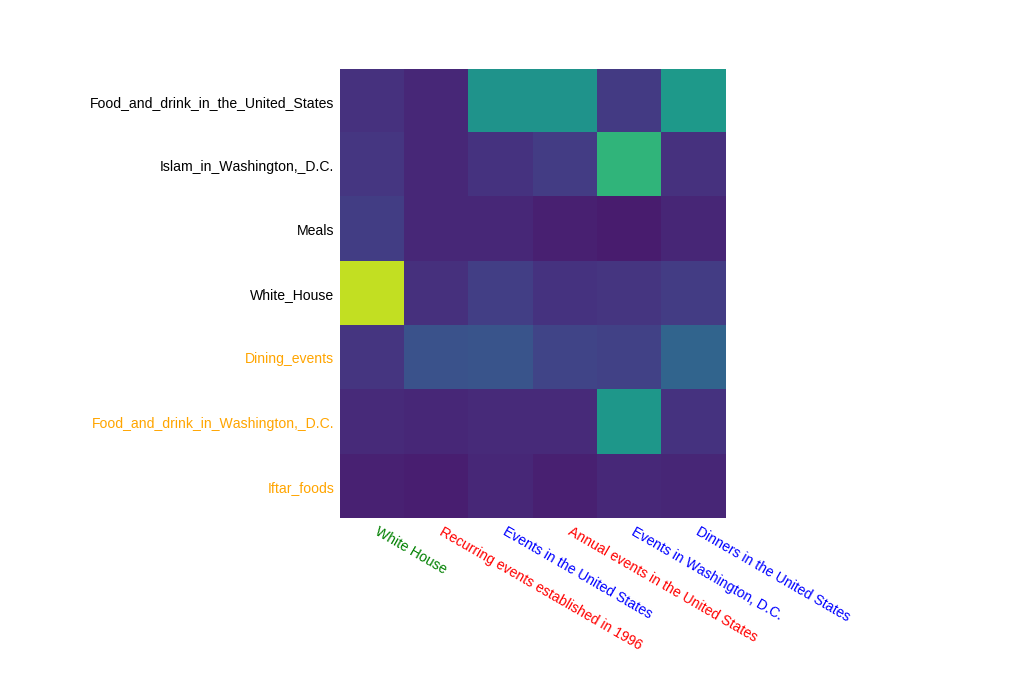} {\small White\_House\_Iftar\_dinner use American English date June 2017 use mdy dates date June 2017 The White House Iftar dinner is an annual reception held at the White House and hosted by the President of the United States U S President and the First Lady of the United States First Lady to celebrate the Muslim month of Ramadan The annual tradition started in 1996 when Hillary Clinton hosted a Ramadan Eid al Fitr Eid celebration Iftar dinner The modern iteration of the reception is attended by prominent members of the Muslim American community including politicians community leaders and students Thomas Jefferson held the first White House dinner with a Muslim while hosting Sidi Soliman Mellimelli an envoy of Beylik of Tunis on December 9 1805 during the First Barbary War lt ref gt cite web last Shellnutt first Kate date August 4 2011 title Thomas Jefferson held first White House Ramadan celebration website IIP Digital publisher blog chron com url http blog chron com believeitornot 2011 08 thoma...} \\

\hline

\begin{tabular}{p{0.7\linewidth}}
    \textit{"People's Democratic Party Turkey Politicians"}: sensible but \underline{not informative} as there is another very similar label in gold set that could have been generated \\
    \hline
    \textit{"MEPs for Turkey 2014-19"}: sensible and informative \\ 
\end{tabular} 
&
\includegraphics[width=\linewidth]{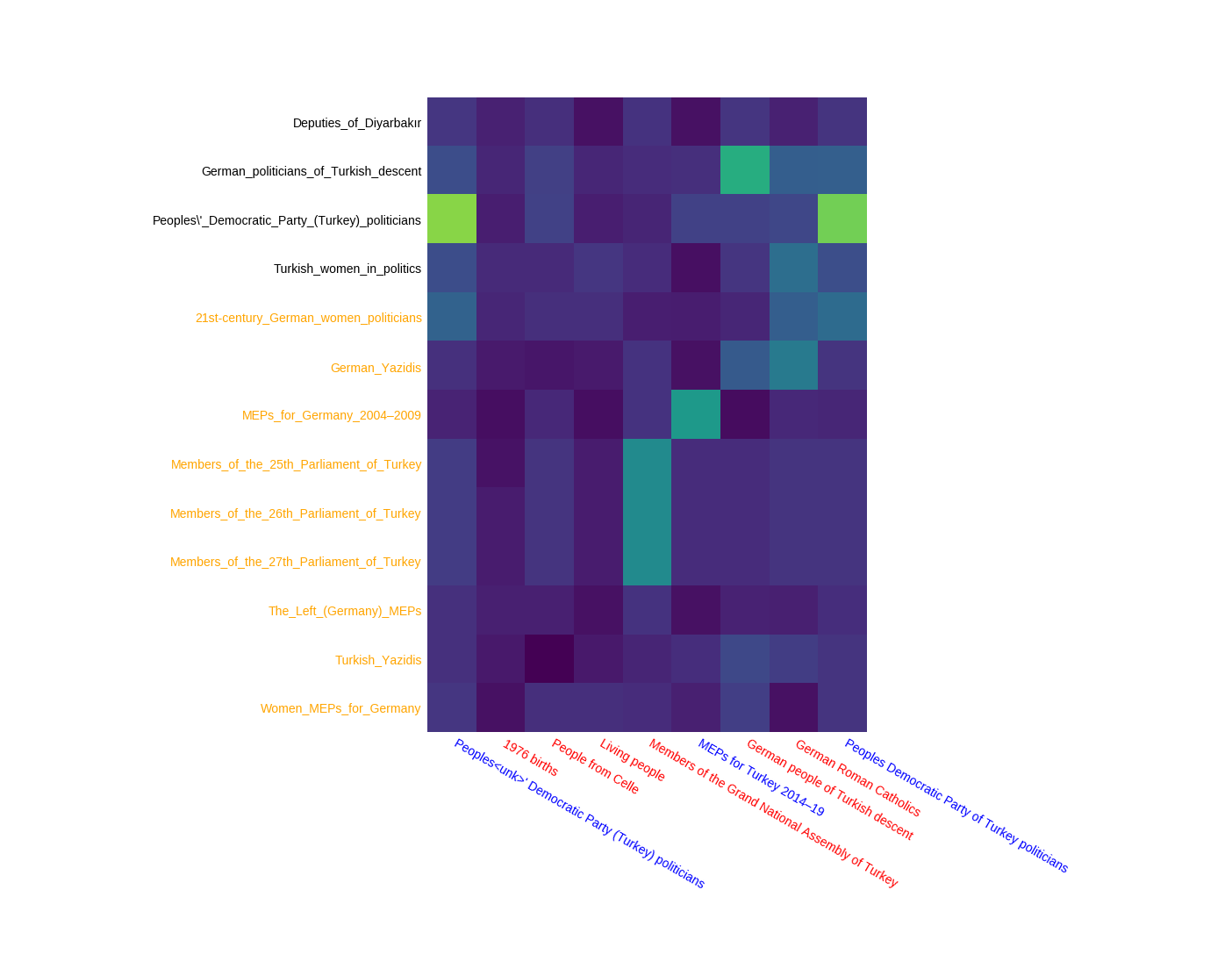} {\small Feleknas\_Uca Use dmy dates date October 2013 Infobox officeholder name Feleknas Uca office Grand National Assembly of Turkey Composition Member of the Grand National Assembly honorific suffix Member of Parliament Turkey MP image Feleknas Uca jpg constituency Diyarbak r electoral district Diyarbak r June 2015 Turkish general election June 2015 November 2015 Turkish general election Nov 2015 lt br gt Batman electoral district Batman 2018 Turkish general election 2018 signature signature\_alt party Peoples Democratic Party Turkey Peoples Democratic Party lt br gt lt br gt otherparty Party of Democratic Socialism Germany Party of Democratic Socialism 1999 2007 lt br gt The Left Germany Die Linke 2007 2009 office1 Member of the European Parliament for Germany birth\_date Birth date and age 1976 09 17 birth\_place Celle Lower Saxony West Germany death\_date lt Death date and age YYYY MM DD YYYY MM DD gt death\_place resting\_place nationality alma\_mater occupation website awards image\_size 220px t...} \\

\hline

\begin{tabular}{p{0.7\linewidth}}
    \textit{"Valhalla Enterteinment films"}: sensible and informative as there is another very similar label in gold set that could have been generated \\
\end{tabular} 
&
\includegraphics[width=\linewidth]{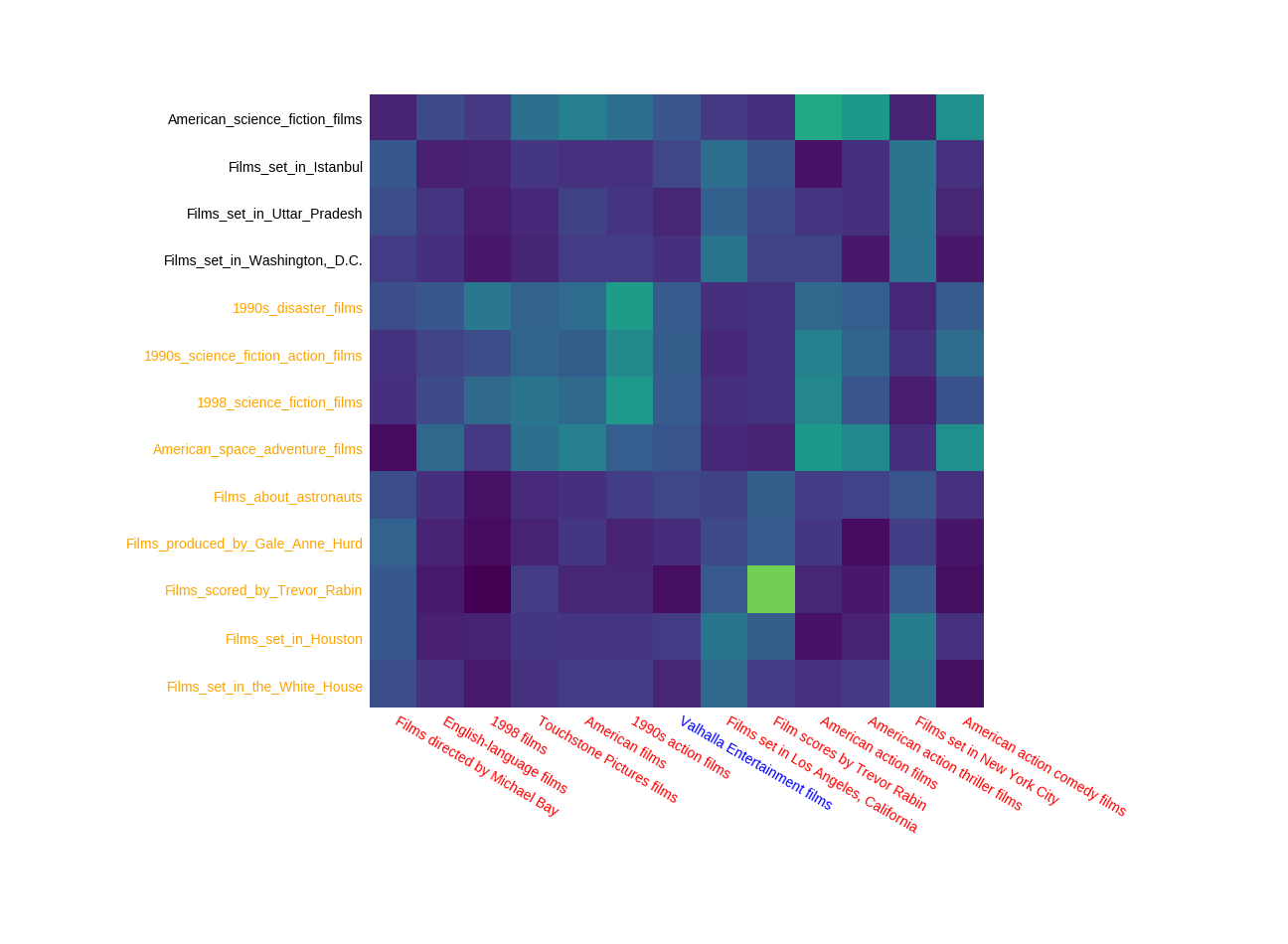} {\small Armageddon\_(1998\_film) use mdy dates date June 2012 Infobox film name Armageddon image Armageddon poster06 jpg alt caption Theatrical release poster director Michael Bay producer Plainlist Jerry Bruckheimer Gale Anne Hurd Michael Bay screenplay Plainlist Jonathan Hensleigh J J Abrams story Plainlist Robert Roy Pool Jonathan Hensleigh starring plainlist Bruce Willis Billy Bob Thornton Liv Tyler Ben Affleck Will Patton Peter Stormare Keith David Steve Buscemi narrator lt Used in documentaries only gt music Plainlist Trevor Rabin cinematography John Schwartzman editing Plainlist Mark Goldblatt Chris Lebenzon Glen Scantlebury studio Plainlist Touchstone Pictures Jerry Bruckheimer Films Valhalla Entertainment Valhalla Motion Pictures distributor Buena Vista Pictures released Film date 1998 07 01 runtime 151 minutes lt Theatrical runtime 150 20 gt lt ref gt cite web url https bbfc co uk releases armageddon 1970 6 title ARMAGEDDON 12 work British Board of Film Classification date July 7 1998 ...} \\

\hline

\begin{tabular}{p{0.7\linewidth}}
    \textit{"Bulgaria Under-20 international footballers"}: sensible and informative \\
\end{tabular} 
&
\includegraphics[width=\linewidth]{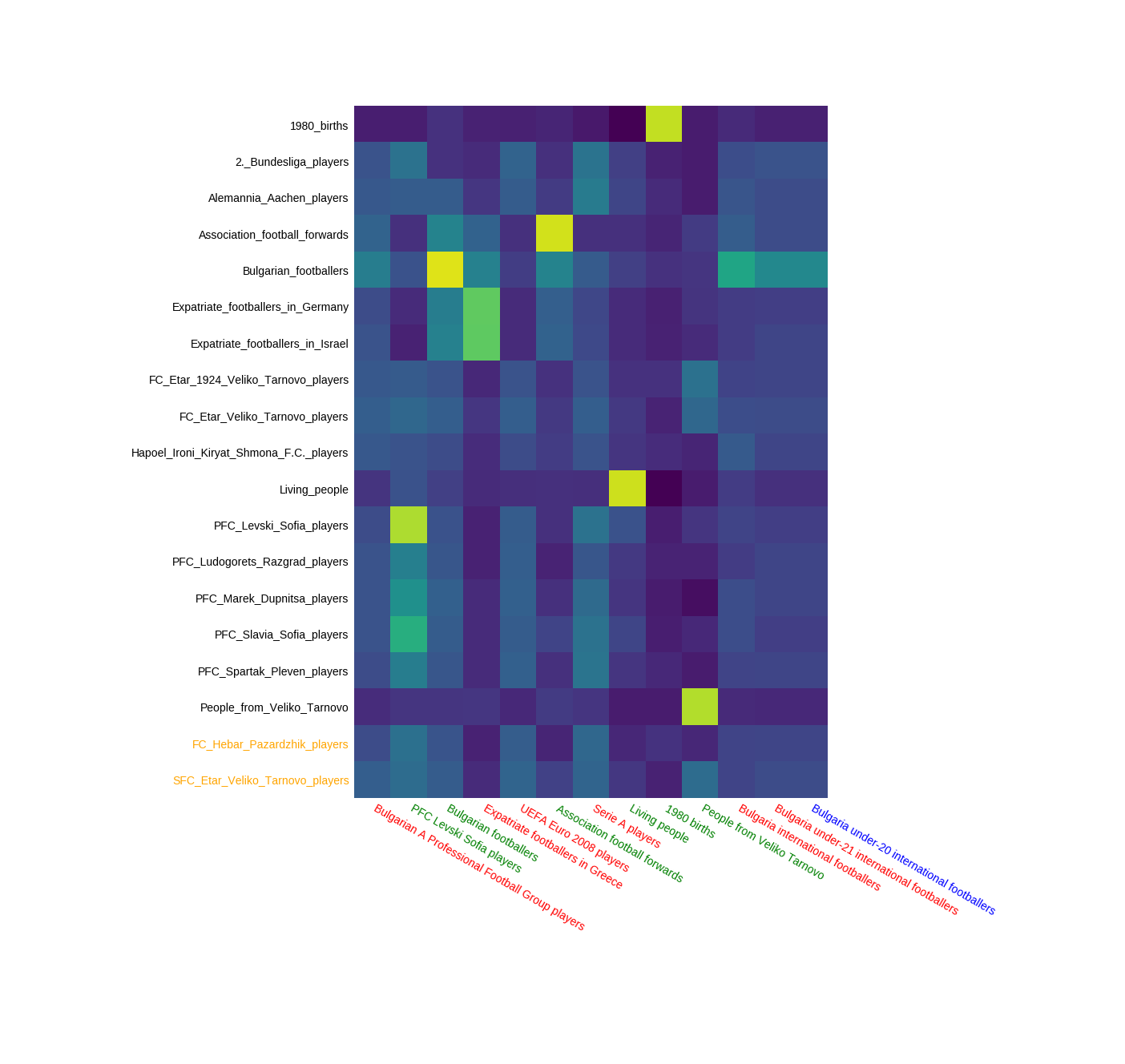} {\small Todor\_Kolev\_(footballer,\_born\_1980) Other people Todor Kolev Use dmy dates date August 2012 Infobox football biography name Todor Kolev image Kolev todor jpg caption Kolev playing for Ludogorets in 2011 fullname Todor Aleksandrov Kolev birth\_date Birth date and age 1980 2 8 df y birth\_place Veliko Tarnovo Bulgaria height convert 1 86 m ftin 0 abbr on currentclub SFC Etar Veliko Tarnovo Etar II Etar Veliko Tarnovo II clubnumber 10 position Forward association football Forward youthyears1 youthclubs1 F C Etar Etar Veliko Tarnovo years1 1997 1999 clubs1 F C Etar Etar Veliko Tarnovo caps1 goals1 years2 1999 2005 clubs2 PFC Levski Sofia Levski Sofia caps2 55 goals2 16 years3 2000 2002 clubs3 PFC Spartak Pleven Spartak Pleven loan caps3 49 goals3 57 years4 2005 clubs4 PFC Marek Dupnitsa Marek Dupnitsa loan caps4 4 goals4 1 years5 2005 2007 clubs5 PFC Slavia Sofia Slavia Sofia caps5 55 goals5 32 years6 2007 2008 clubs6 Alemannia Aachen caps6 20 goals6 5 years7 2008 2010 clubs7 PFC Slavia Sofi...} \\

\hline

&
\includegraphics[width=\linewidth]{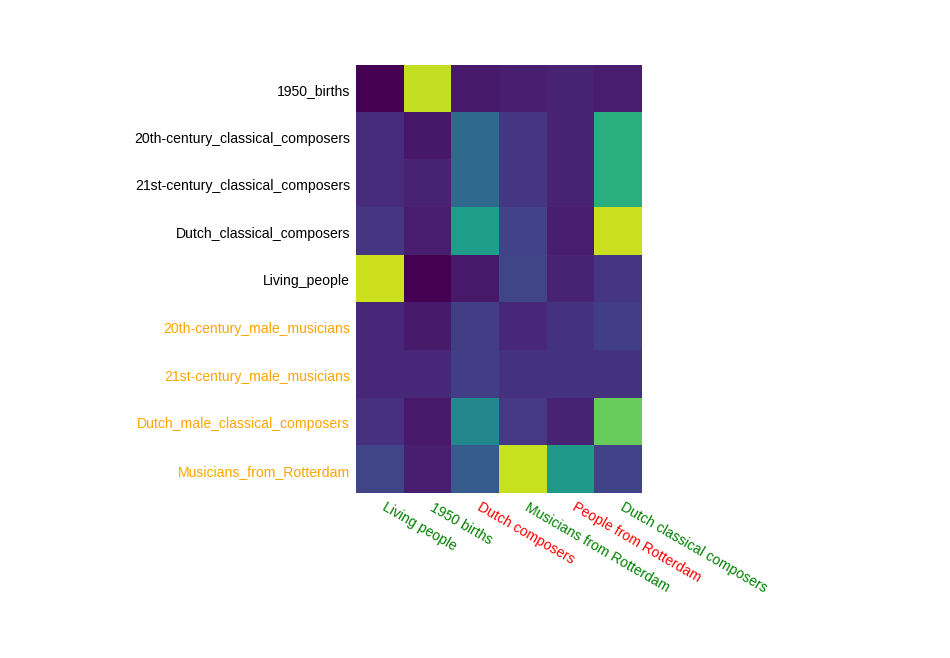} {\small John\_Borstlap John Borstlap 4 November 1950 Rotterdam is a Dutch composer lt ref gt cite book title Entartete Musik publisher Emanuel Overbeeke amp Leo Samama url https books google com id NydqmVZUhlEC amp pg PA175 amp lpg PA175 amp dq john borstlap v onepage amp q john 20borstlap amp f false isbn 9789053567159 year 2004 lt ref gt and author on cultural subjects related to music and the visual arts He claims to be rooted in German musical traditions and is a proponent of a revival of tonal and classical traditions} \\

\hline
  
\begin{tabular}{p{0.7\linewidth}}
    \textit{"Artists from Changzhou"}: sensible and informative \\
    \hline
    \textit{"Qianlong people"}: sensible and informative \\ 
\end{tabular} 
&
\includegraphics[width=\linewidth]{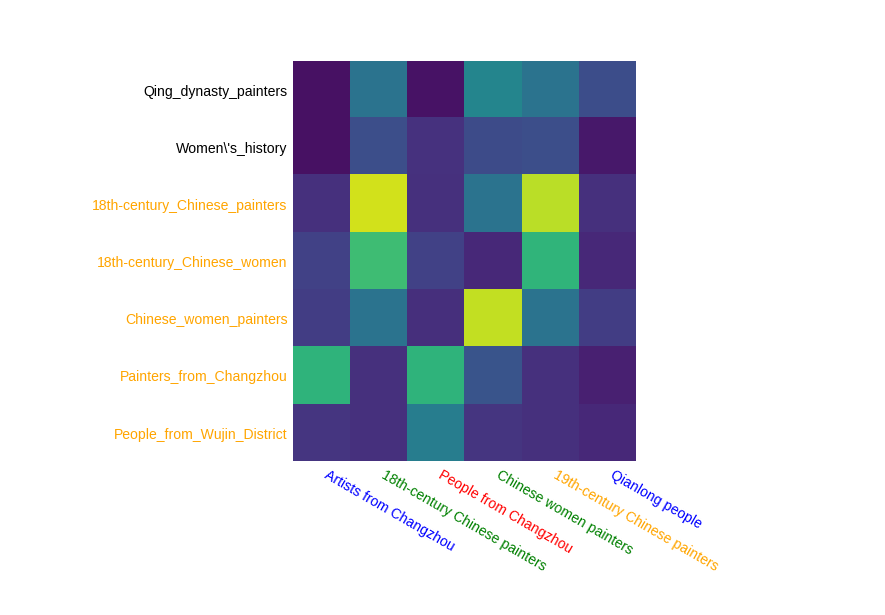} {\small Yun\_Bing Infobox artist name Yun Bing native\_name native\_name\_lang zh birth\_place Wujin District Changzhou known\_for notable\_works Hairpin Scroll 1735 1796 lt br gt Quiet Provisions of the Studio 1735 1796 style Bird and flower painting quot Boneless quot technique movement spouse Mao Hongtiao module Infobox Chinese child yes t s p Y n B ng w Y n Ping altname Qingyu c2 linktext p2 Q ngy w2 Ch ing y patrons memorials Yun Bing zh c dates unknown courtesy names Qingyu zh c and Haoru zh c was a Chinese painter during the Qianlong era She is well known for her bird and flower painting s executing the quot boneless quot technique and became the most famed of the Yun family s female artists lt ref name lu gt cite title trans title Discussion of the achievements of the influential family near the mound the Yun clan language Chinese author Lu Haiyang journal Changzhou gong xueyuan xuebao shekeban volume 31 issue 1 date 2013 pages 1 7 lt ref gt} \\
  
\hline
\end{longtable}

\end{document}